\documentclass[preprint,authoryear,11pt]{elsarticle}
\usepackage[margin=1.5in]{geometry}

\usepackage{url}
\usepackage{graphics}
\usepackage{tabularx}
\usepackage{multirow}
\usepackage{rotating}
\usepackage{alltt}
\usepackage{amsmath}
\usepackage{amssymb}
\usepackage{longtable}
\usepackage{pdflscape}
\usepackage{natbib}
\usepackage{framed}
\usepackage{comment}
\usepackage[doublespacing]{setspace}

\journal{Decision Support Systems}

\begin{document}

\begin{frontmatter}

\title{Predicting Process Behaviour using Deep Learning}

\author[mun]{Joerg Evermann\corref{cor1}}
\cortext[cor1]{Corresponding author}
\ead{jevermann@mun.ca}

\author[dfki,unisb]{Jana-Rebecca Rehse}
\author[dfki,unisb]{Peter Fettke}

\address[mun]{Memorial University of Newfoundland, St. John's, NL, Canada}
\address[dfki]{German Research Center for Artificial Intelligence, Saarbr\"ucken, Germany}
\address[unisb]{Saarland University, Saarbr\"ucken, Germany}

\begin{abstract}
Predicting business process behaviour is an important aspect of business process management. Motivated by research in natural language processing, this paper describes an application of deep learning with recurrent neural networks to the problem of predicting the next event in a business process. This is both a novel method in process prediction, which has largely relied on explicit process models, and also a novel application of deep learning methods. The approach is evaluated on two real datasets and our results surpass the state-of-the-art in prediction precision.
\end{abstract}


\end{frontmatter}

\section{Introduction}
\label{sec:intro}

Being able to predict the future behaviour of a business process is an important business capability \citep{DBLP:conf/ifip8/HouyFLAK10}. As an application of predictive analytics in business process management, process prediction exploits data on past process instances to make predictions about current ones \citep{Breukeretal:2016}. Example use cases are customer service agents responding to inquiries about the remaining time until a case is resolved, production managers predicting the completion time of a production process for better planning and higher utilization, or case managers identifying likely compliance violations to mitigate business risk. 

We present a novel approach to predicting the next process event using deep learning. While the term ''deep learning'' has only recently become a popular research topic, it is essentially an application of neural networks and thus looks back on a long history of research \citep{DBLP:journals/nn/Schmidhuber15}. Recent innovations both in algorithms, allowing novel architectures of neural networks, and computing hardware, especially GPU processing, have led to a resurgence in interest for neural networks and popularized the term ''deep learning'' \citep{LeCunetal:2015}. Our approach is motivated by applications of neural networks to Natural Language Processing (NLP), more specifically the prediction of the next word in a sentence \citep{DBLP:conf/icml/SutskeverMH11,DBLP:journals/corr/Graves13,DBLP:journals/corr/ZarembaSV14}. By interpreting process event logs as text, process traces as sentences, and process events as words, these techniques can be applied to predict future process events. The contribution of our research is threefold:

\begin{enumerate}
\item We improve on the state-of-the-art in process event prediction. Our results show our method has considerably better precision on next-event prediction.
\item We demonstrate that an explicit process model is not necessary for prediction. Deep learning models, where the process structure is only implicitly reflected, can perform as well as explicit process models.
\item We contribute to process management in general by showcasing the useful application of an artificial intelligence approach, illustrating that business process management can benefit from the application of smart approaches.
\end{enumerate}

Our research is located at the intersection of business process management, in particular process mining, and artificial intelligence (AI) and machine learning. We bring together historic process data with an AI learning technology to leverage real-time case management, opening new perspectives into process execution, monitoring, and analysis. Extending existing solutions to novel problems (''exaptation'') is a recognized and valid way to make a contribution in design science \citep{DBLP:journals/misq/GregorH13}, which is the research approach we apply here. We not only provide a new approach, rooted in AI, to predicting the next process event, but also give a proof-of-concept regarding its feasibility and experimentally explore its efficiency and effectiveness, thus making a valuable contribution to the field of ''Smart BPM''.

This paper is a significant extension over earlier work \citep{Evermannetal:2016}, adding more advanced neural network cells, separation of training and validation samples for cross-validation to prevent overfitting, empirical assessment of the effect of different neural network parameters, prediction not only of next events but of case remainders, interpretation and visualization of neural network states, encoding of timing information, and an extended discussion of the similarities and differences between natural language processing and process event prediction.


\section{Related Work}
\label{sec:related}

Process prediction covers an array of different techniques, objectives, and data sources. It extends process mining from a post-hoc analysis method to operational decision support \citep{vanderAalst:2010}. Most existing process prediction research focuses on prediction of process outcomes, primarily the remaining time to completion, rather than prediction of the next event in a process, as we do here. Only five approaches are concerned with predicting the next event \citep{Breukeretal:2016,DBLP:conf/dis/CeciLFCM14,DBLP:journals/kais/LakshmananSDUK15,DBLP:conf/sgai/LeGN12,DBLP:journals/kais/UnuvarLD16}, all of which use an explicit model representation such as a state-transition, HMM (hidden Markov models), or PFA (probabilistic finite automatons) model. 

The MSA approach by \citet{DBLP:conf/sgai/LeGN12} considers each trace prefix as a state in a state-transition matrix. From the observed prefixes and their next events, a state transition matrix is built. When a running case has reached a state not contained in the state-transition matrix, its similarity to observed traces is computed using string edit distance. The prediction is made from the most similar observed case. Evaluating the approach on two datasets from a telecommunications company, \citet{DBLP:conf/sgai/LeGN12} report accuracies in predicting the next event of up to 25\% and 70\% for their two datasets.

The approach described by \citet{DBLP:journals/kais/LakshmananSDUK15} and \citet{DBLP:journals/kais/UnuvarLD16} consists of five steps. First, a process model is mined from existing logs. For each XOR split in the model, a decision tree is mined from case data. These trees are then used to compute the state transition probabilities for a HMM that is specific to the running case that is to be predicted. This HMM is then used to predict the probabilities of the following event. The approach is evaluated on simulated data. After training the decision trees for the HMM on one half of the traces (training set), each trace in the test set is cut into a prefix and postfix at a random point. For each prefix, the log-likelihood of observing the corresponding postfix is reported.

The approach described by \citet{DBLP:conf/dis/CeciLFCM14} uses sequence mining to identify frequent trace prefixes. For each prefix, a regression model is trained to predict remaining time to completion and a decision tree is trained to predict the next event. The algorithm identifies the appropriate prefix of the running case to choose the regression and decision tree model for predicting remaining completion time and the next event. The experimental evaluation uses two datasets, yielding prediction precision values for the next event of approximately 65\% on one dataset and approximately 50\% on the other, depending on the type of decision tree used and the frequency threshold for prefixes.

RegPFA \citep{Breukeretal:2016} uses a probabilistic finite automaton (PFA) instead of a HMM, allowing the future hidden state to be a function of both the previous hidden state and the previous observed event (which itself is a probabilistic consequence of the previous hidden state). RegPFA uses an EM algorithm to estimate the model parameters of the PFA. The evaluation uses data from the 2012 and 2013 BPI Challenges \citep{bpic2012dataset,bpic2013dataset2,bpic2013dataset1}.

Our approach has the same objective as these five related works, but differs in terms of method and process representation. Deep Learning in the form of a recurrent neural network (RNN) is used to predict the next events, using event sequences and associated resource information. Processes are only implicitly represented within the RNN, rather than in an explicit state-transition, PFA or HM model. Because of the non-linear transformations used in neural networks, they are particularly suited to model non-linear relationships. In contrast, many of the existing methods are based on linear assumptions (e.g. regression trees). Thus, neural networks require less restrictive assumptions. Moreover, because models are built implicitly rather than explicitly, their performance does not depend on any ex-ante assumptions about the form of the model. Overall, our method constitutes an innovative new approach to process prediction. 

\section{Deep Learning for Process Prediction}
\label{sec:nn}

\subsection{Introduction}

A neural network is a special form of an acyclic computational graph \citep{DBLP:journals/nn/Schmidhuber15}. It consists of a layer of input cells, one or more layers of ''hidden'' cells, and a layer of output cells. Cells in each layer are connected by weighted connections to cells in the previous and following layers, allowing for different network architectures. Each cell's output is a function of the weighted sum of its inputs. A simple network architecture is a fully connected network of cells using sigmoid activation functions that form the hidden layer.

\vspace{-25pt}
\begin{align*}
a_j^l = \sigma \left( \sum_i w_i^{l,j} a_i^{l-1} + b_j^l \right) \quad \text{where} \quad \sigma(x) = \frac{1}{1+exp(x)}
\end{align*}

\noindent Here, $a_j^l$ is the output (''activation'') of cell $j$ in layer $l$, $w_j^{l,i}$ is the weight of the connection from cell $i$ on layer $l-1$ to cell $j$ on layer $l$, $a_i^{l-1}$ is the output of cell $i$ on layer $l-1$ and $b_j^l$ is the ''bias'' of cell $j$ on layer $l$.

A neural network is a supervised learning technique where the output of the neural net is compared to a target by means of a loss function. The type of output layer cell and the loss function are often chosen jointly for their computational properties with respect to backpropagation. A typical combination for categorical targets is a softmax layer with a cross-entropy loss function $H$:

\vspace{-25pt}
\begin{alignat*}{2}
y_i = softmax(a)_i &= \frac{exp(a_i)}{\sum_j exp(a_j)} && \qquad H_{y\prime}(y) = - \sum_i y_i\prime log(y_i)
\end{alignat*}

\noindent Here, $y\prime$ are the target values and $y$ are the network outputs, computed in turn from the output activations $a_i$ of the next to last network layer. Gradients of the parameters $w_j^{l,i}, b_i^{l}$ with respect to the loss function are computed using backpropagation and parameters are adjusted using variants of gradient descent algorithms. 

\subsection{Recurrent Neural Networks (RNN)}

\begin{figure}
\vspace{-35pt}
\centerline{\includegraphics[width=\textwidth]{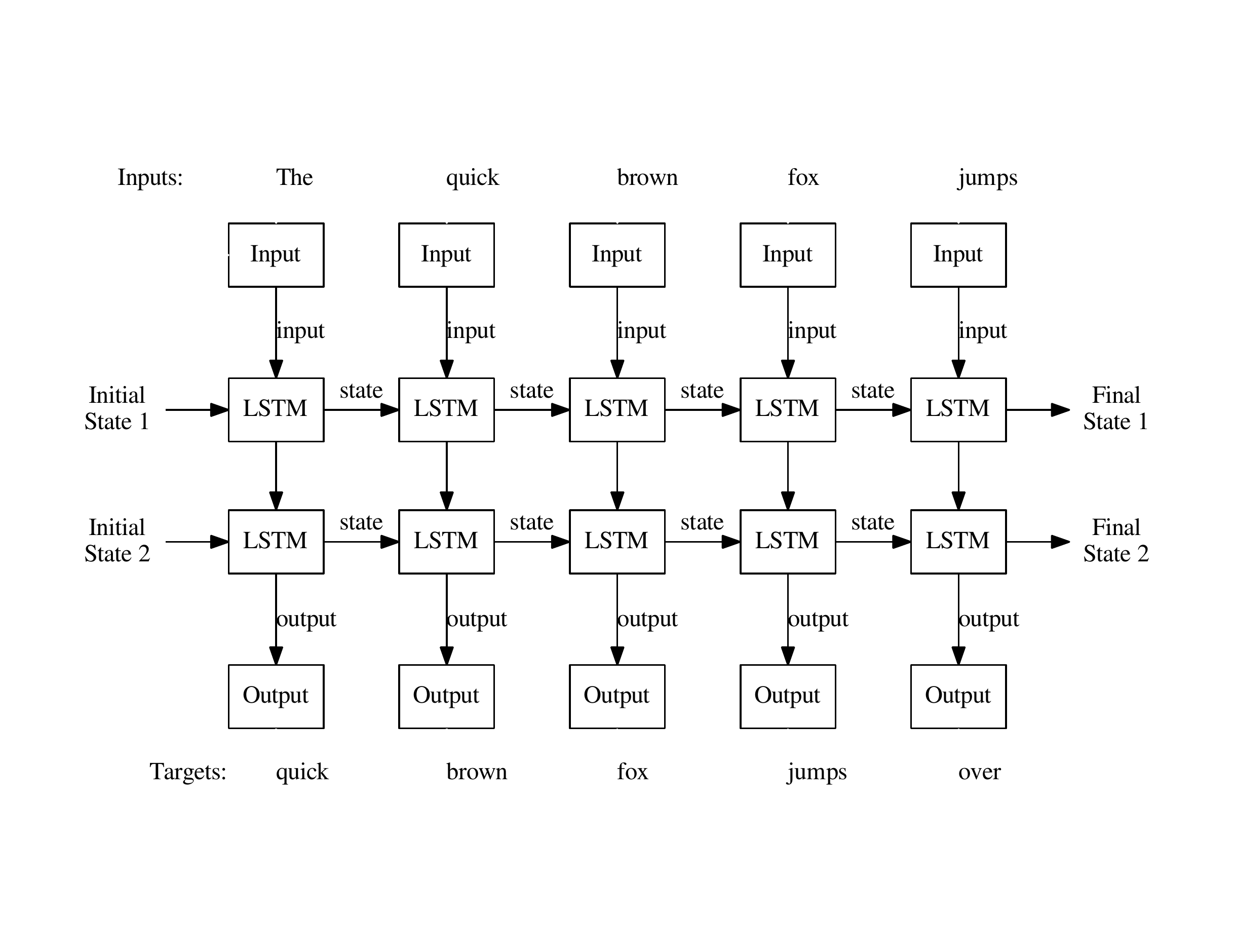}}
\vspace{-45pt}
\caption{RNN architecture with single hidden layer of LSTM cells, unrolled five steps}
\label{fig:rnnarchitecture}
\end{figure}

Recurrent neural networks (RNN) are a neural network architecture popularized by work in natural language processing (NLP). NLP has moved away from explicit language models to statistical methods, specifically to RNN \citep[for example,][]{mikolov2011extensions,DBLP:conf/icml/SutskeverMH11,DBLP:journals/corr/Graves13,DBLP:series/sci/2012-385,DBLP:journals/corr/ZarembaSV14}. In a RNN, each cell's output is not only connected to the following layer but each cell also feeds back information into itself, allowing it to maintain ''state'' over time. To make this tractable within an acyclic computational graph and backpropagation, the recurrent network cells are ''unrolled'', that is, copies of it are produced for time $t, t-1, t-2, \ldots$. The state output of the RNN cell for time $t-1$ is state input to the cell for time $t$. In general, $t$ need not represent time, but can index any sequence. Depending on how long one wishes to maintain state for, fewer or more cells are unrolled. Fig.~\ref{fig:rnnarchitecture} shows an RNN architecture with an input layer, an output layer and two hidden layers that are unrolled five steps \citep{DBLP:journals/corr/Graves13}. Each layer (each box in Fig.~\ref{fig:rnnarchitecture}) in turn consists of multiple input, output, or hidden cells that are not individually shown in Fig.~\ref{fig:rnnarchitecture}.

A typical NLP application trains the RNN on sequences of input words to predict the next word, e.g. to provide word suggestions for user input. As shown in Fig~\ref{fig:rnnarchitecture}, the target words are simply the input words shifted by one position, so that for each input word the following word is the target to be predicted. The ability to maintain state or context means that, for example, when given the input "fox" in Fig.~\ref{fig:rnnarchitecture} and predicting the correct target (the next word) "jumps", the RNN can make use of the previous words "The quick brown" to improve prediction performance. 

In this work, we apply a recurrent neural network to the problem of predicting the next event in a process from a sequence of observed events. In implementing process prediction using RNN, our main idea is \emph{to view an event log as a text, traces in the log as sentences, and events in a trace as analogous to words in a sentence}. 

\subsection{Long Short Term Memory (LSTM)}
\label{sec:lstm}

Recurrent networks with simple sigmoid cells have unsatisfactory performance for long time or sequence distances, leading to the development of long short term memory (LSTM) cells \citep{DBLP:journals/neco/HochreiterS97}, defined by the vector equations in Fig~\ref{fig:lstm}. An LSTM cell accepts $c_{t-1}$ and $h_{t-1}$ as state information from the prior unrolled cell on the same level, and $x_t$ as input from cells on the previous layer. In turn, it passes $c_t$ and $h_t$ as new state information to the subsequent unrolled cell on the same level and also provides $h_t$ as output to the next layer.

\begin{figure}
\noindent
\begin{minipage}{0.5\textwidth}
\begin{align}
f_t &= \sigma \left( W_f \cdot \left[ h_{t-1}, x_t \right] + b_f \right) \label{eq1} \\
i_t &= \sigma \left( W_i \cdot \left[ h_{t-1}, x_t \right] + b_i \right)  \label{eq2} \\ 
\bar{c}_t &= \tanh \left( W_c \cdot \left[ h_{t-1}, x_t \right] + b_c \right) \label{eq3}
\end{align}
\end{minipage}\begin{minipage}{0.5\textwidth}
\begin{align}
c_t &= f_t \times c_{t-1} + i_t \times \bar{c}_t  \label{eq4} \\
o_t &= \sigma \left( W_o \cdot \left[ h_{t-1}, x_t \right] + b_o \right)  \label{eq5} \\
h_t &= o_t \times \tanh ( c_t )  \label{eq6}
\end{align}
\end{minipage}
\caption{Definition of LSTM cell}
\label{fig:lstm}
\end{figure}

The variables $c, x, h, f, i, o$ in Fig.~\ref{fig:lstm} are vectors in $\mathbb{R}^m$; the $W$ and $b$ (weights and biases) are ''trainable'' parameters of suitable dimensions. The functions $\tanh(\ldots)$ and $\sigma(\ldots)$ are applied elementwise, $[ \ldots ] $ represents vector concatenation. Equations~\ref{eq1}--\ref{eq6} describe $m$ individual LSTM cells, each operating on a single input from $\mathbb{R}$. The number of individual cells on each layer, i.e. the dimensionality $m$, can be freely chosen. 

The intuition behind these definitions is as follows. Eq.~\ref{eq1} represents the ''forget gate'' that determines, based on the inputs $x_t$ and $h_{t-1}$, which part of the state to forget. Note that the prior state $c_{t-1}$ is multiplied by $f_t$ in Eq.~\ref{eq4} to derive the new state. With $\sigma(\ldots)$ in Eq.~\ref{eq1} yielding values between 0 and 1, some information in $c_{t-1}$ will, to some degree, be ''forgotten''. Eq.~\ref{eq2} and \ref{eq3} determine how the inputs $x_t$ and $h_{t-1}$ contribute to the updated cell state. Eq.~\ref{eq2} represents the ''input gate'' and determines which values of the state to update; some of the $i_t$ will be close to zero, others close to one. Eq.~\ref{eq3} computes new candidate values $\bar{c}$, which are multiplied with $i_t$ in Eq.~\ref{eq4} when computing the new state $c_t$. Eq.~\ref{eq4} is the actual state update equation. The left term represents the ''forgetting'' of some prior state information while the right term represents the addition of new information to the cell state. Eq.~\ref{eq5} represents the ''output gate''. It determines which values of the cell state are provided as output to the following layer and subsequent unrolled cell. It is multiplied in Eq.~\ref{eq6} with the $\tanh$ of the cell state to produce the final output $h_t$. 

To further improve the performance of LSTM cells, \citet{DBLP:conf/ijcnn/GersS00} introduce the idea of ''peepholes'', which allow the forget and input gates to look (''peep'') at the prior state $c_{t-1}$, and which allows the output gate to also look (''peep'') at the current state $c_{t}$, with the equations in Fig.~\ref{fig:peepholes} replacing those in Fig.~\ref{fig:lstm}. While there exist other variants to this basic LSTM cell, \citet{DBLP:journals/corr/GreffSKSS15} conclude after an experimental evaluation that ''the most commonly used LSTM architecture, performs reasonably well on various datasets and using any of eight possible modifications does not significantly improve the LSTM performance.''

\begin{figure}
\begin{align*}
f_t &= \sigma \left( W_f \cdot \left[ c_{t-1}, h_{t-1}, x_t \right] + b_f \right) \\
i_t &= \sigma \left( W_i \cdot \left[ c_{t-1}, h_{t-1}, x_t \right] + b_i \right) \\ 
o_t &= \sigma \left( W_o \cdot \left[ c_t, h_{t-1}, x_t \right] + b_o \right) \\
\end{align*}
\vspace{-35pt}
\caption{Definition of peepholes for LSTM cells}
\label{fig:peepholes}
\end{figure}

\subsection{Word Embeddings}

As neural networks operate on real-valued data, words or any other categorical input must be appropriately encoded. One option for this is to use ''one-hot'' encodings, but this leads to input vectors and neural net layers whose size (the dimensionality $m$ in Sec.~\ref{sec:lstm}) is the number of distinct words or categories. Word embeddings avoid this problem and allows arbitrary input vector and neural network sizes. Words are mapped into an n-dimensional ''embedding'' space $\mathbb{R}^{v \times m}$ using an ''embedding matrix'', which is essentially a look-up matrix of dimensions $v \times m$ where $v$ is the size of the vocabulary and $m$ is the chosen dimensionality of the embedding space and hence the size of each LSTM hidden layer (cf. Section~\ref{sec:lstm}). The input layer in Fig.~\ref{fig:rnnarchitecture} is an embedding lookup function yields a numeric vector from $\mathbb{R}^m$ for each word in the vocabulary, which forms the input $x_t$ in Eq.~\ref{eq1}, \ref{eq2} and \ref{eq5} for the first LSTM layer. The embedding matrix is also a trainable parameter that is learned during training. The output layer produces a probability distribution over the words in the vocabulary. Each output ''box'' in Fig.~\ref{fig:rnnarchitecture} represents a softmax layer with $v$ individual cells. The word with the highest probability is selected as the predicted word and compared to the target.

\subsection{Language and Process}
\label{sec:prediction}

Referring to the XES standard \citep{xes}, we define events jointly by the name of the executing activity (e.g. ''Pay Invoice'', ''Repair Widget''), the lifecycle transition (e.g. ''Start'', ''Complete'', ''Schedule'') and, for some experimental settings, the organizational resource, role or group (e.g. ''Jane Doe'', ''Tester'', ''Accounting''). An example of an event is ''Repair Widget---Start---Jane Doe''. Our ''vocabulary'' is the set of unique events of this form contained in a log.

As natural language is constrained by grammatical and morphological rules, such as noun and verb agreement for plurals in English, process event sequences are determined or constrained by an underlying process logic, for example by business rules based on case data. Just as linguistic rules and constraints are not explicitly captured in NLP deep-learning approaches \citep{mikolov2011extensions,DBLP:conf/icml/SutskeverMH11,DBLP:series/sci/2012-385,DBLP:journals/corr/Graves13,DBLP:journals/corr/ZarembaSV14} but are learned by the neural network, the process constraints and rules also need not be explicitly represented but can be learned. 

However, while there are many similarities between natural language and process traces, there are important differences. First, the size of the vocabulary in process prediction is much smaller than the size of a natural language vocabulary. Second, the length of a trace can far exceed the typical sentence length in natural language. Together, these two differences result in fewer possible prediction targets (vocabulary size or number of unique process event types) and more information to predict from (sentence or trace prefix length), suggesting that this approach may be able to achieve better results than word prediction in NLP. Third, in contrast to natural language, processes contain activities with temporally overlapping execution. But even in that case, individual lifecycle events are strictly ordered, e.g. ''A---Start, B---Start, A---Complete, B---Complete'', as defined by the \citet{xes}: ''An event that occurs in a log $\ldots$ before another event that is related to the same trace, shall be assumed to have occurred before that other event.'' However, the possibly arbitrary temporal sequence of events of parallel activities across different traces can make the prediction task more difficult, especially when there are no underlying regularities or dependencies imposed by other process characteristics such as resources or case attributes.

\begin{figure}
\begin{minipage}{0.48\textwidth}
\centering
\includegraphics[width=\textwidth]{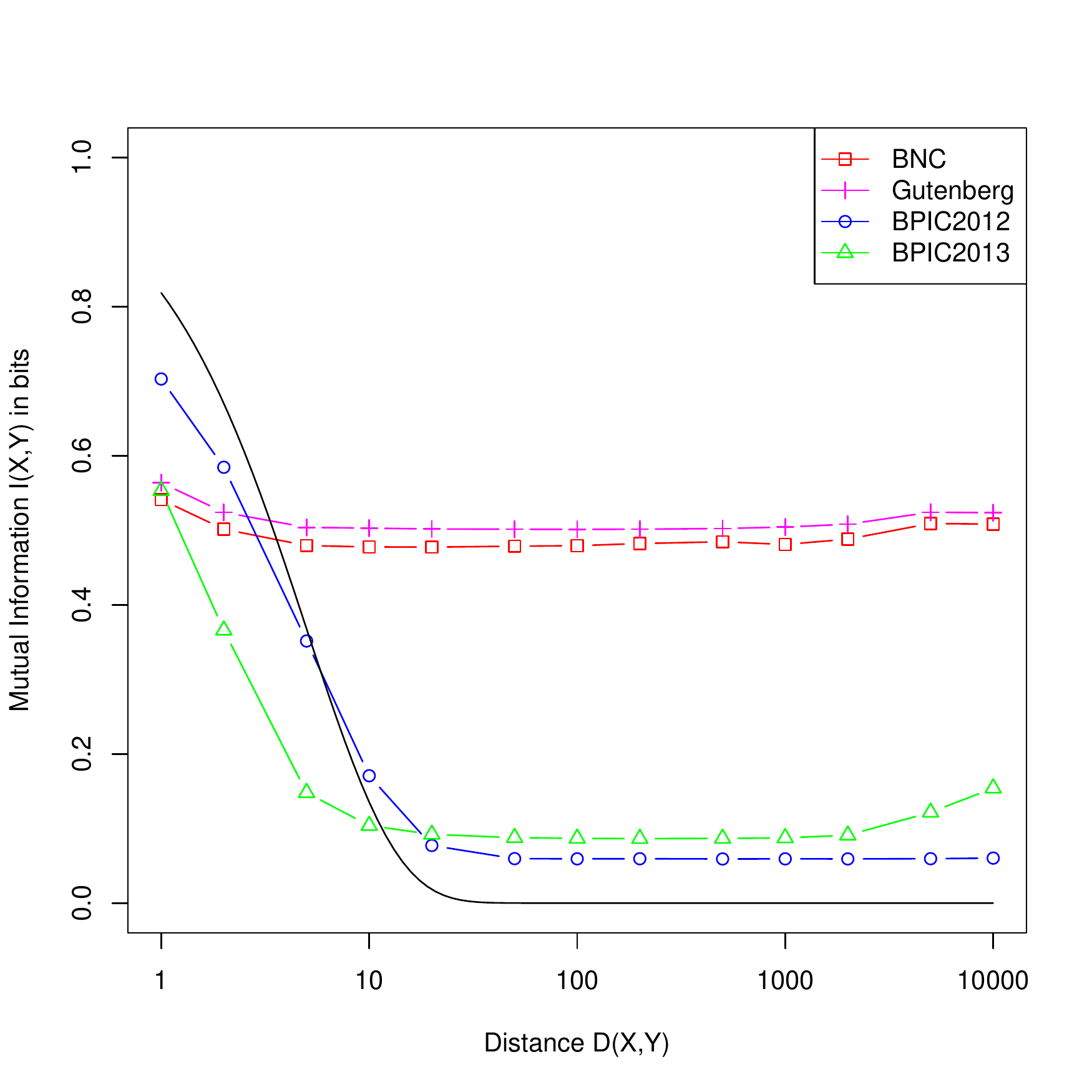}
\caption{Mutual information of process event logs and English language text by separation distance (solid line is $I = \exp(-d/5)$)}
\label{fig:mutualinformation}
\end{minipage}
\quad
\begin{minipage}{0.48\textwidth}
\centering
\includegraphics[width=\textwidth]{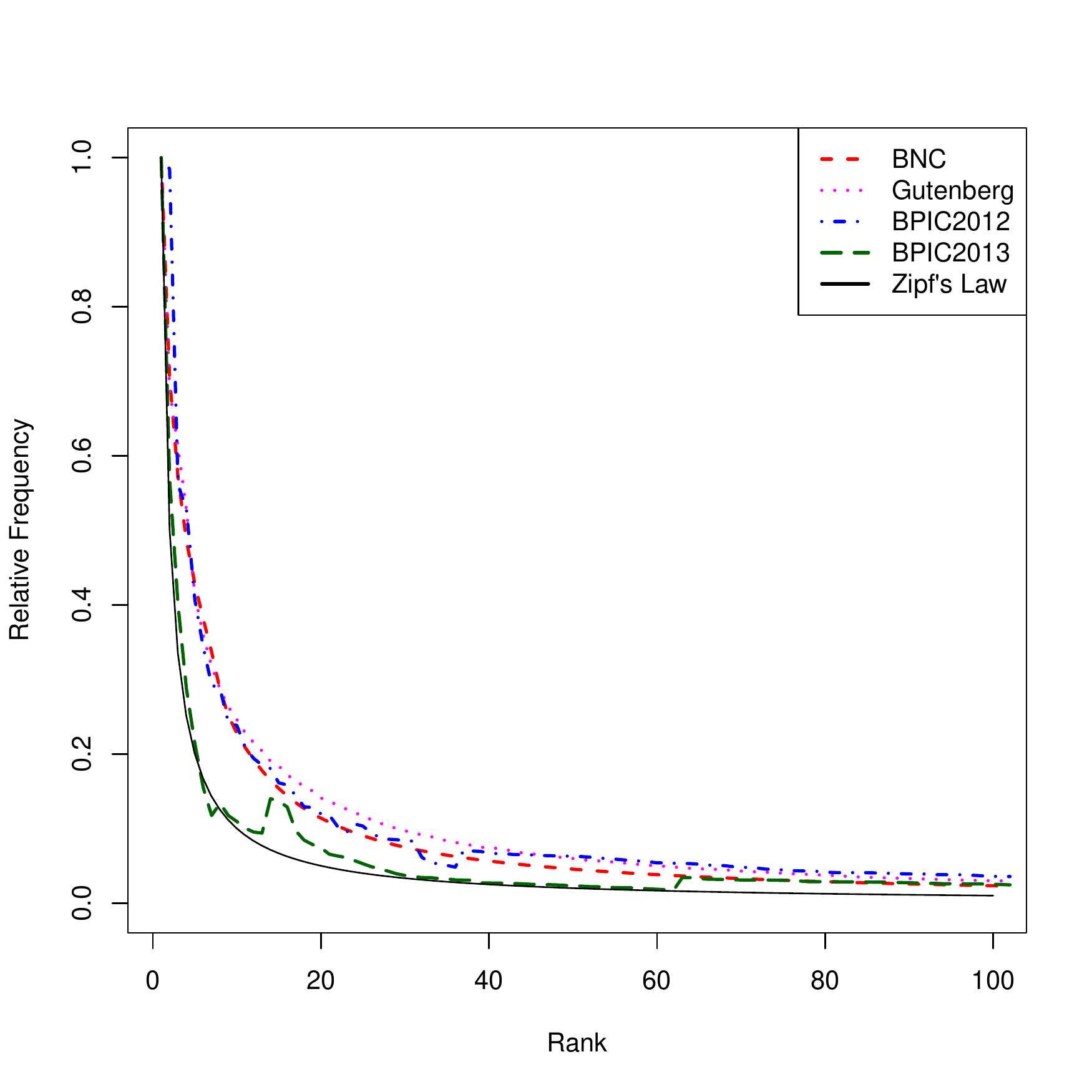}
\caption{Relative frequency of terms by frequency rank for process event logs and English language text (solid line is Zipf's law)}
\label{fig:zipf}
\end{minipage}
\end{figure}

To empirically examine the similarity of process event logs and natural language text, we focus on two characteristics. First, the mutual information $I$ of two words in natural language declines with their distance $d$ according to a power law ($I \propto d^{-k}$), rather than an exponential manner ($I \propto \exp{-d/k}$). Markov models cannot model this form of dependency, while RNN can \citep{2016arXiv160606737L}. We compare the mutual information in our datasets against that of two natural language text corpora, the British National Corpus\footnote{\url{http://www.natcorp.ox.ac.uk/}} and an extract from Project Gutenberg\footnote{\url{https://www.gutenberg.org/}} (Fig.~\ref{fig:mutualinformation}). As our event logs exhibit exponential decline, an RNN is not strictly necessary as they can be modelled by both Markov-based models and RNNs. This invites a comparison between the RNN and Markov-based approach, which we provide by comparing our method to the PFA method by \citet{Breukeretal:2016} (Sec.~\ref{sec:experiment}). 

Second, we examine whether Zipf's law \citep{george1949zipf} holds for the event logs. Zipf's law states that the frequency of words in a natural language corpus is inversely proportional to the frequency rank of that word. Figure~\ref{fig:zipf} shows that our event logs also adhere to Zipf's law and are, in this characteristic, similar to natural language. 

In summary, given that there are both similarities and differences, a proof-of-concept implementation to demonstrate feasibility of applying NLP methods to prcoess prediction, and an experimental evaluation to demonstrate effectiveness and performance are clearly required.




\section{Implementation}
\label{sec:implementation}

A number of software frameworks for deep-learning have become available recently \citep{bahrampour2015comparative}. We implement our approach using Tensorflow as it provides a suitable level of abstraction, provides RNN specific functionality, and can be used on high-performance parallel, cluster, and GPU computing platforms. Code, data and complete results are available from the corresponding author\footnote{\url{http://joerg.evermann.ca/software.html}}.

Our network features an architecture as in Fig.~\ref{fig:rnnarchitecture} with two hidden RNN layers using LSTM cells. This architecture still affords many choices, in particular the dimensionality of the embedding space and the number of unrolled steps.

The dimensionality of the embedding space must be guided by the vocabulary size (the number of unique event types). A larger dimensionality allows better separation of words in that space, which likely leads to better predictive performance, but at the cost of computational effort. Our baseline in the experimental evaluation (Sec.~\ref{sec:evaluation}) is $125$, and we explore the effect of varying this parameter on predictive performance. The number of ''unrolled'' steps indicates the length of the sequence of words across which state information is maintained, independent of vocabulary size. A larger number of unrolled steps allows the network to take earlier process events into account when predicting the following process event. This, too, is at the expense of computational effort. Our baseline for the experimental evaluation is $20$, and we explore the effect of varying this parameter on predictive performance. 

In addition to these two parameters, a number of general neural network options are parameterized (Table~\ref{tab:techparameters}). \citet{DBLP:journals/corr/GreffSKSS15} show that these parameters are significantly less important in determining RNN performance than network size and are largely independent in their effect on RNN performance. Trainable parameters are initialized using a uniform random distribution over $[-0.1, 0.1]$. Training proceeds in batches of size $20$. For each batch, the backpropagation algorithm computes the mean gradients for all parameters. Training of the net proceeds in ''epochs''. Each epoch trains the net on the entire event log. Subsequent epochs maintain the weights $W$ and biases $b$ learned from the previous epoch but reinitialize the states for each layer and then train the net again on the entire event log. The net was trained for $100$ epochs. The learning rate is reduced from $1$ by a factor of $0.75$ each epoch after the 50th. Dropout is a technique to prevent overfitting and improve learning by randomly temporarily removing a cell from the network \citep{DBLP:journals/jmlr/SrivastavaHKSS14}. Dropout probability for each cell is $0.2$.

\begin{table}
\centering
\small

\begin{tabular}{|l|r|} \hline
\emph{Parameter} & \emph{Value} \\ \hline
Initialization scale & 0.10 \\
Batch size & 20 \\ 
Maximum gradient norm & 5.00 \\
Number of training epochs & 100 \\
Number of epochs with full base learning rate & 50 \\
Base learning rate & 1.00 \\
Learning rate decay & 0.75 \\
Dropout probability & 0.20 \\
LSTM forget bias & 0.10 \\ \hline
\end{tabular}
\caption{RNN Parameters}
\label{tab:techparameters}

\end{table}

\section{Experimental Method}
\label{sec:evaluation}

\subsection{Data}

To provide a compelling evaluation of our approach, it should be compared to the state-of-the-art in next event prediction. Of the related work discussed in Section~\ref{sec:related}, only \citet{Breukeretal:2016} make an implementation publicly available and use publicly available data, demonstrating ''open research'' \citep{DBLP:journals/bise/AalstBH16}. We contacted all authors of the remaining papers twice, but did not receive software or data to use for comparative evaluation. 

We use the same datasets as \citet{Breukeretal:2016}. The BPI Challenge 2012 dataset \citep{bpic2012dataset} is a real dataset from a loan application process in a Dutch financial institute with 13087 traces. It can be separated into three subprocesses concerning the application itself (A), the offer (O) and the work item (W) belonging to the application. The BPI Challenge 2013 datasets \citep{bpic2013dataset2,bpic2013dataset1} are real datasets from IT incident and problem management processes at Volvo Belgium with 7553 and 2300 traces, respectively.

In addition to separating the BPI 2012 data set by sub-process as done by \citet{Breukeretal:2016}, we also use the combined dataset. While \citet{Breukeretal:2016} use only activity completion events for the BPI 2012 dataset, we also test our approach on all events (including the lifecycle transitions ''Start'', ''Schedule'' and ''Complete''). However, only the ''W'' subset has events other than completion events. Furthermore, we include an experimental condition where we extract the activity name and lifecycle transition, and combine this with the name of the resource associated with the event (cf. Sec.~\ref{sec:prediction}). We simply concatenate the two character strings to form the composite word. This creates a larger vocabulary which increases the prediction difficulty, but also provides more information to the training algorithm. It also allows prediction of not only the activity of the next event but also the resource associated with the next event. Because the number of distinct resources in the BPI 2013 datasets is very large, we use the organizational group instead of organizational resource. Table~\ref{tab:datasets} shows characteristics of the datasets.

\begin{table}
\centering
\small

\begin{tabular}{|l|r|r|r|} \hline
\multirow{2}{*}{Dataset} & \multicolumn{2}{c|}{\parbox[c][][c]{1.7in}{Number of unique event types (''vocabulary size'')}} & \multirow{2}{*}{\parbox[c][][c]{.7in}{Number of events}} \\
                         &  & w/ res. info & \\ \hline
BPI2013.Incidents 						& \parbox{.85in}{\raggedleft 14} 	& 3133 	&  65533 \\ 
BPI2013.Problems 						& 8 		& 64 	&   9011 \\ 
BPI2012 (completion events) 				& 24 	& 877	& 164506 \\ 
BPI2012 (all events) 					& 37		& 1349	& 262200 \\ 
BPI2012.W (completion events) 			& 7 		& 264	&  72413 \\ 
BPI2012.W (all events) 					& 7		& 736	& 169507 \\ 
BPI2012.A 							& 11		& 302	&  60849 \\ 
BPI2012.O					 			& 8 		& 313	&  31244 \\ 
\hline
\end{tabular}
\caption{Characteristics of datasets used in experimental evaluation. Event numbers for partial BPI2012 logs do not add up to that of corresponding complete log due to end-of-case marker events added to each trace.}
\label{tab:datasets}
\end{table}

The published datasets are transformed using XSL transformations to extract traces, events, and resource information in a suitable format. 

\subsection{Evaluation Method}

There are three random influences on prediction performance. First, the trainable parameters are initialized randomly and different initial values may lead to different outcomes (e.g. convergence to local optima). Second, the traces in a log are in a particular order but this is an arbitrary order. Because the LSTM cells maintain state, the order in which traces are used to train the network may have an effect on the outcome. Finally, the selection of the training dataset itself has an influence, and the results obtained with a particular sample may not generalize to others. Both BPI 2012 and 2013 datasets are \emph{sampled} from running systems. 

To address these stochastic influences, we perform 10-fold cross-validation \citep{Hastieetal:09}. Comparing the prediction performance on an independent validation sample to the prediction performance on the training sample assesses the generalizability to similar datasets. Specifically, it examines whether the RNN \emph{overfits} the training sample, i.e. exploits idiosyncrasies in the training sample, and what type of performance can realistically be expected on a dataset with similar characteristics. Note that \cite{Breukeretal:2016} do not cross-validate the results of their stochastic EM-based approach; a fair comparison is therefore to our training results. 

We report the prediction precision, defined as the proportion of correct predictions of all predictions made. We report mean and standard deviation of the training precision, as well as the mean and standard deviation of the precision achieved on the validation fold, across all 10 folds. 

\section{Experimental Results}
\label{sec:experiment}

\begin{figure}
\centering \par
\includegraphics[width=\textwidth]{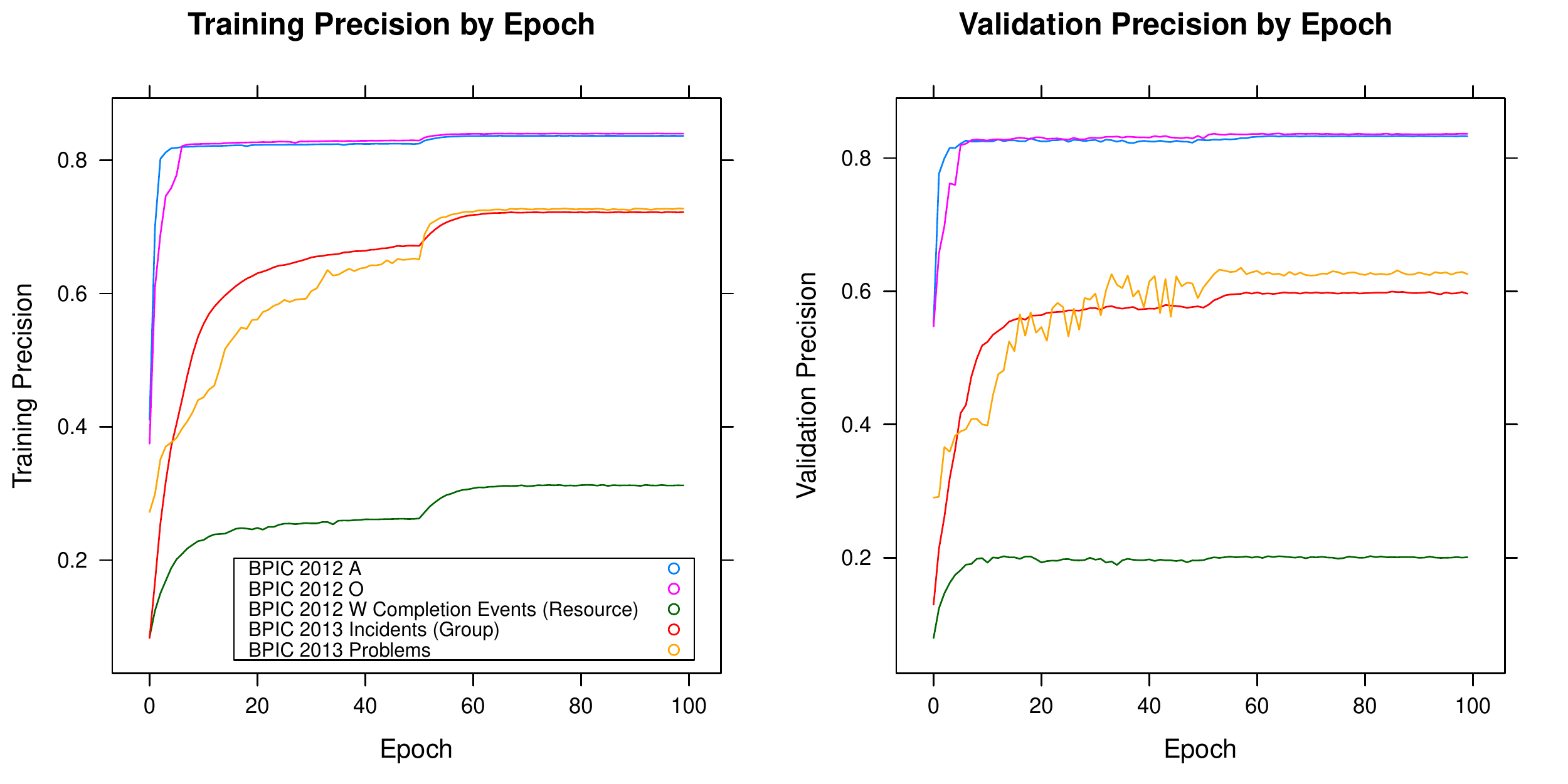}
\caption{Training precision by training epoch for selected datasets (mean over 10 training folds)}
\label{fig:precision}
\end{figure}

We train the RNN for 100 epochs on each training dataset. Fig.~\ref{fig:precision} plots training and validation precision for each epoch for a selection of our datasets (averaged across all 10 training folds). The plot shows that 100 training epochs are sufficient for optimal and stable results. The small BPI 2012 A and O datasets converge quickly to a high precision, whereas this occurs more gradually for the BPI 2013 datasets. Moreover, the BPI 2012 A, BPI 2012 O, and BPI 2013 Problem datasets with relatively small vocabularies converge faster than the BPI 2012 W and BPI 2013 Incident datasets that include resource or organizational group information and consequently have a larger vocabulary. There is a significant improvement in precision when the training rate is adjusted in epoch 50, suggesting that a dynamic training rate helps to prevent suboptimal convergence. The second panel in Fig.~\ref{fig:precision} shows that validation precision generally follows the behaviour of training precision at a lower level.

\subsection{Predicting the Next Event}

\begin{table}
\centering
\small

\begin{tabular}{|l|r|r|r|r|r|} \hline
\multirow{3}{*}{Dataset} & \multirow{3}{0.8in}{\centering Precision in \citep{Breukeretal:2016}} & \multicolumn{2}{c|}{\multirow{2}{0.8in}{\centering Training Precision}} & \multicolumn{2}{c|}{\multirow{2}{0.8in}{\centering Validation Precision}} \\
 & & \multicolumn{2}{c|}{} & \multicolumn{2}{c|}{} \\
 & & Mean & SD & Mean & SD \\ \hline

BPI2013.Incidents 				& .714 & .757 & .011 & .735 & .044 \\ 
BPI2013.Problems 				& .690 & .727 & .037 & .628 & .086 \\ 
BPI2012 (completion events) 		&      & .796 & .001 & .788 & .006 \\ 
BPI2012 (all events) 			&      & .864 & .001 & .859 & .005 \\ 
BPI2012.W (completion events) 	& .719 & .676 & .002 & .658 & .020 \\ 
BPI2012.W (all events) 			&      & .839 & .001 & .832 & .010 \\ 
BPI2012.A 				 	& .801 & .837 & .001 & .832 & .010 \\ 
BPI2012.O 					& .811 & .841 & .001 & .836 & .010 \\ 
\hline 
\end{tabular}
\caption{Results and comparison to \citep{Breukeretal:2016}}
\label{tab:results}
\end{table}

Table~\ref{tab:results} shows our results and a comparison to the best result presented by \citet{Breukeretal:2016}. The table shows that our RNN approach surpasses the performance of probabilistic finate automatons on many datasets. For the BPI 2013 Incidents data, the BPI 2012 A, and the BPI 2012 O subsets, our training and validation precision values are above the best values reported by \citet{Breukeretal:2016}. For the BPI 2013 Problems dataset, our approach offers better training precision, but the validation precision is lower than that reported by \citet{Breukeretal:2016}, and on the BPI 2012 W dataset, our approach performs poorly. Overall, this is in line with the expectations based on mutual information (Sec.~\ref{sec:prediction}) which showed that both PFA and RNN should be able to model our process data equally well.

Comparing the training and validation precision shows that little overfitting occurs, with the validation precision generally within $\pm 0.05$ of the training precision. The standard deviations for the validation precision are an order of magnitude above those for the training sample because of the smaller size of the validation sample (one tenth of the training sample size).

Table~\ref{tab:results} shows many results with validation precision close to or in excess of $0.8$. While we have no comparison to the state-of-the-art on these datasets by \citet{Breukeretal:2016}, this level of precision is encouraging for practical applications. It is also generally above the levels reported in related works (see Section~\ref{sec:related}), although this must be interpreted with caution due to different datasets with possibly very different characteristics, such as number of unique events, trace length, or the patterns due to the structure of the underlying, generating process.

Prediction is significantly better for the BPI 2012 W dataset with all events compared to the same dataset with completion events only as it exploits the regularity that after every ''Start'' event for an activity, the corresponding ''Complete'' event for that activity follows.

\subsection{The Effect of Resource Information}

Including the organizational resource or group in the predicting data provides additional information that should lead to improved prediction precision. Including this information in both predictor and predictand allows prediction not only of the activity of the next process event but also of the resource associated with the next process event. However, this comes at the cost of a larger vocabulary and should lead to lower prediction precision. We explore both options, and the results in Table~\ref{tab:resultsresources2} are generally as expected.

\begin{table}
\centering
\small

\begin{tabular}{|l|r|r|r|r||r|r|r|r|} \hline
\multirow{4}{*}{Dataset} & \multicolumn{4}{c||}{Predictand only} & \multicolumn{4}{c|}{Predictor \& Predictand} \\
& \multicolumn{2}{c|}{\multirow{2}{0.5in}{Training Precision}} & \multicolumn{2}{c||}{\multirow{2}{0.5in}{Validation Precision}} & \multicolumn{2}{c|}{\multirow{2}{0.5in}{Training Precision}} & \multicolumn{2}{c|}{\multirow{2}{0.5in}{Validation Precision}} \\
 & \multicolumn{2}{c|}{}  & \multicolumn{2}{c||}{} & \multicolumn{2}{c|}{}  & \multicolumn{2}{c|}{}  \\
 & Mean & SD & Mean & SD & Mean & SD & Mean & SD \\ \hline
BPI2013.Incidents 				& .832 & .034 & .761 & .049 & .731 & .010 & .595 & .061 \\		
BPI2013.Problems  				& .776 & .081 & .616 & .117 & .764 & .034 & .517 & .043 \\	  	
BPI2012 (completion) 		& .833 & .004 & .811 & .023 & .622 & .005 & .594 & .010 \\ 		
BPI2012 (all events) 			& .880 & .007 & .866 & .008 & .685 & .007 & .663 & .015 \\		
BPI2012.W (completion) 	& .745 & .013 & .693 & .019 & .313 & .012 & .208 & .023 \\ 		
BPI2012.W (all events) 			& .865 & .012 & .845 & .012 & .609 & .018 & .559 & .018 \\ 		
BPI2012.A 					& .846 & .006 & .826 & .007 & .758 & .010 & .709 & .008 \\ 		
BPI2012.O 					& .870 & .015 & .833 & .015 & .703 & .026 & .590 & .020 \\		
\hline
\end{tabular}
\caption{Precision when including resources or organizational groups}
\label{tab:resultsresources2}
\end{table}



Including the organizational group or resource information only in the predictor (Table~\ref{tab:resultsresources2}) improves the training precision for all datasets, by up to $0.07$, compared to the baseline in Table~\ref{tab:results}. However, the validation precision does not follow the same pattern, improving only for some datasets, and even then improving to a lesser extent as the training precision. This suggests that including resource information for prediction may be useful but also runs the danger of overfitting the model.




Including the organizational resource or group in both the predictor and predictand affects the prediction precision for the datasets in different ways. Training precision for the BPI 2013 Incident decreases, but increases for the BPI 2013 Problem dataset. Validation precision drops significantly for both datasets. Training and validation precision for all BPI 2012 datasets also drop significantly. An extreme example is the BPI 2012 W dataset with completion events only, where training precision drops to $.313$ and validation precision to $.208$. Here, the resources assigned to events do not follow any underlying regularity, impairing prediction performance. In contrast, for the BPI2012.W with all events, the resource of the completion event is the same as of the start event, which leads to improved performance.

\subsection{Predicting Duration of Activities}

An RNN can also be used to predict the duration of activities. For this, we quantize the temporal extent of a trace in minutes and, for each minute, encode the current activity in the input. For example, if activity A occurs for 3 minutes, the input consists of the sequence AAA. Only the BPI 2012 W dataset includes both start and completion events to allow determination of the duration of activities\footnote{This kind of encoding is sensible only when activities do not temporally overlap, which is the case for this dataset.}. While it is possible to encode idle time between two activities in the same manner, the BPI 2012 W dataset describes long-running cases, which would have led to very long and monotonous sequences. 

The validation precision for this case is $0.942$, (SD=$0.027$), significantly higher than the $0.832$ reported in Table~\ref{tab:results} for the case with no duration information. This increase in precision is not surprising, as the dataset consists of longer sequences of identical words, thus making prediction of the following word much easier. 


\subsection{The Effect of Embedding Space Dimensionality}

\begin{figure}
\centering \par
\includegraphics[width=\textwidth]{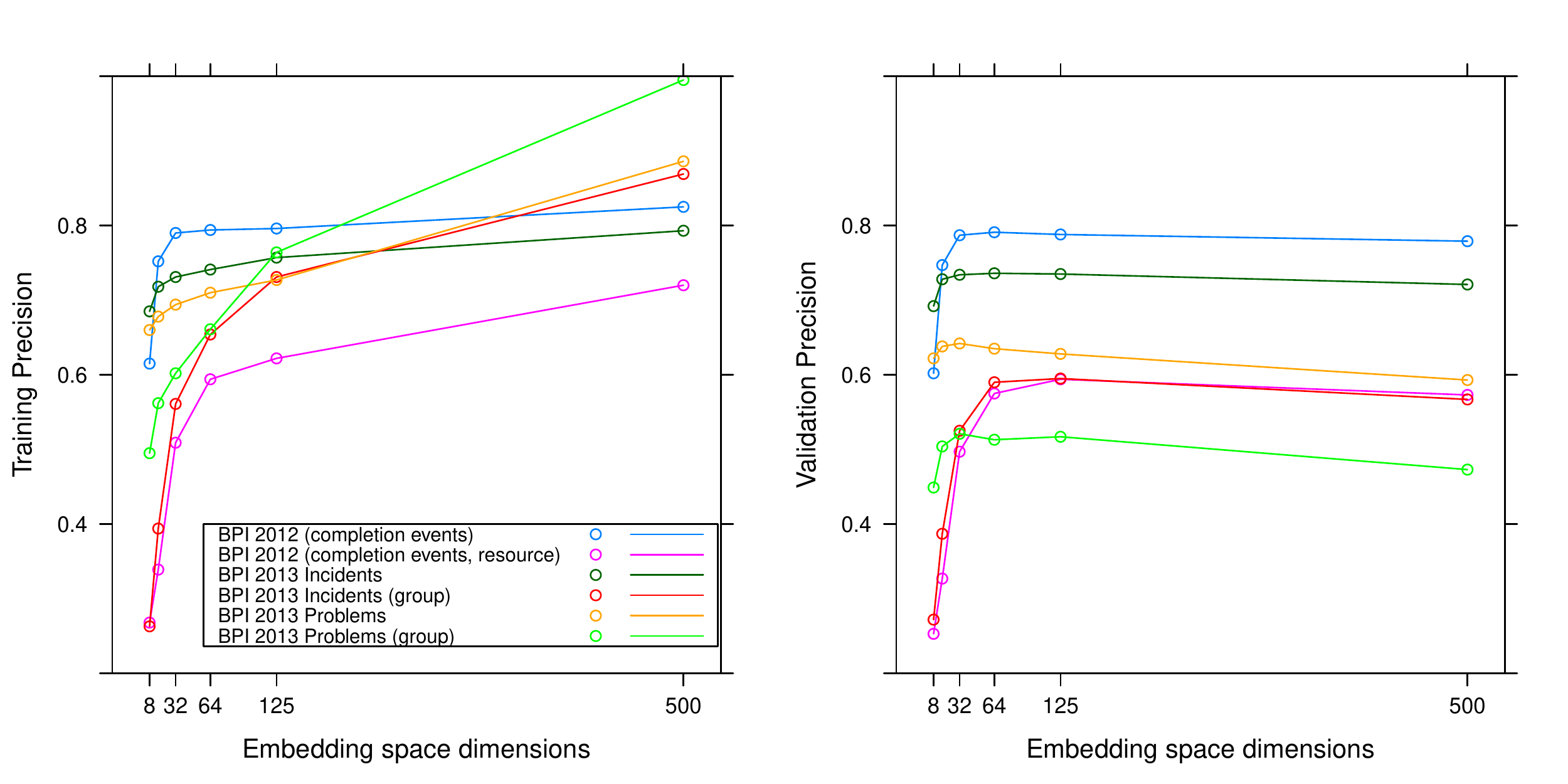}
\caption{Training and validation precision against embedding space dimensions for selected datasets}
\label{fig:accuracybydimension}
\end{figure}

To identify the effect of the dimensionality of the word embedding space on prediction performance, we repeat our experiments using embedding spaces of 500, 64, 32, 16, and 8 dimensions, keeping the other parameters unchanged. Table~\ref{tab:results.dimensionality.validation} shows the validation precision of our datasets for different numbers of dimensions. Fig.~\ref{fig:accuracybydimension} plots training and validation precision for three datasets with and without organizational information.

\begin{table}
\centering
\small

\begin{tabular}{|l|r|r|r|r|r|r|} \hline
\multirow{2}{*}{Dataset} & \multicolumn{6}{c|}{Dimensionality of Embedding Space} \\
 & 500 & 125 & 64 & 32 & 16 & 8 \\ \hline

BPI2013.Incidents 						& .721 & .735 & .736 & .734 & .728 & .692  \\ 
BPI2013.Problems 						& .593 & .628 & .635 & .642 & .638 & .622  \\ 
BPI2012 (completion events) 				& .779 & .788 & .791 & .787 & .747 & .602  \\ 
BPI2012 (all events) 					&      & .859 & .859 & .854 & .772 & .578  \\ 
BPI2012.W (completion events) 			& .638 & .658 & .660 & .661 & .657 & .646  \\ 
BPI2012.W (all events) 					&      & .832 & .833 & .832 & .824 & .733  \\ 
BPI2012.A					 			& .833 & .832 & .833 & .834 & .832 & .799  \\ 
BPI2012.O 					 		& .827 & .836 & .836 & .837 & .836 & .823  \\ 
\hline 
BPI2013.Incidents (with group) 			& .567 & .595 & .590 & .525 & .387 & .272  \\ 
BPI2013.Problems (with group) 			& .473 & .517 & .513 & .521 & .504 & .449  \\  
BPI2012 (completion events, resource) 		& .573 & .594 & .575 & .497 & .327 & .253  \\  
BPI2012 (all events, resource) 			&      & .663 & .618 & .456 & .328 & .188  \\ 
BPI2012.W (completion events, resource) 	& .130 & .208 & .214 & .204 & .174 & .147  \\ 
BPI2012.W (all events, resource) 			&      & .559 & .558 & .495 & .302 & .134 \\  
BPI2012.A (resource) 					& .714 & .709 & .717 & .717 & .688 & .615  \\ 
BPI2012.O (resource)				 	& .577 & .590 & .598 & .588 & .482 & .263  \\ 
\hline

\end{tabular}
\caption{Validation precision for different dimensions of the embedding space. Due to computational requirements, 500 dimensions applied only to selected datasets.}
\label{tab:results.dimensionality.validation}
\end{table}

As expected, reducing the dimensionality of the embedding has, in general, a detrimental effect on the prediction precision. However, this effect is negligible as long as the dimensionality of the embedding space is greater than the size of the vocabulary. This is the case for all datasets that do not include organizational information. For example, neither the BPI 2013 datasets with 14 and 8 event types, nor the BPI 2012 dataset with 24 event types in Fig.~\ref{fig:accuracybydimension} show a marked reduction in training or validation precision when the dimensionality is reduced from 64 to 32 and to 16. However, when the dimensionality is reduced to 8, the reduction in precision is more pronounced for all datasets.

Given the larger vocabulary when including organizational information, the effect of dimensionality is more pronounced at smaller dimensions, as is visible in Fig.~\ref{fig:accuracybydimension}. The BPI 2013 datasets with 3133 and 64 unique combinations of event activity and organizational group, and the BPI 2012 dataset with 877 unique combinations, show a significant reduction in precision when the dimensionality is reduced from 125 to 64. Other datasets with large vocabulary show the same behaviour.

Increasing the dimensionality to 500, done only for selected datasets due to the required computational effort, shows that significant overfitting occurs, especially for datasets with a large vocabulary. As shown in Fig.~\ref{fig:accuracybydimension}, the training precision for the BPI 2012 and BPI 2013 datasets increases significantly whereas the validation precision is not only much lower than the training precision, but decreases with increasing dimensionality. These trends are clear indications of overfitting.

Examining Fig.~\ref{fig:accuracybydimension} and the corresponding data in Table~\ref{tab:results.dimensionality.validation} shows that our chosen baseline with 125 dimensions is close to the optimum validation precision for all datasets and does not suffer from significant overfitting.

\subsection{The Effect of the Number of Unrolled Steps}

\begin{figure}
\centering \par
\includegraphics[width=\textwidth]{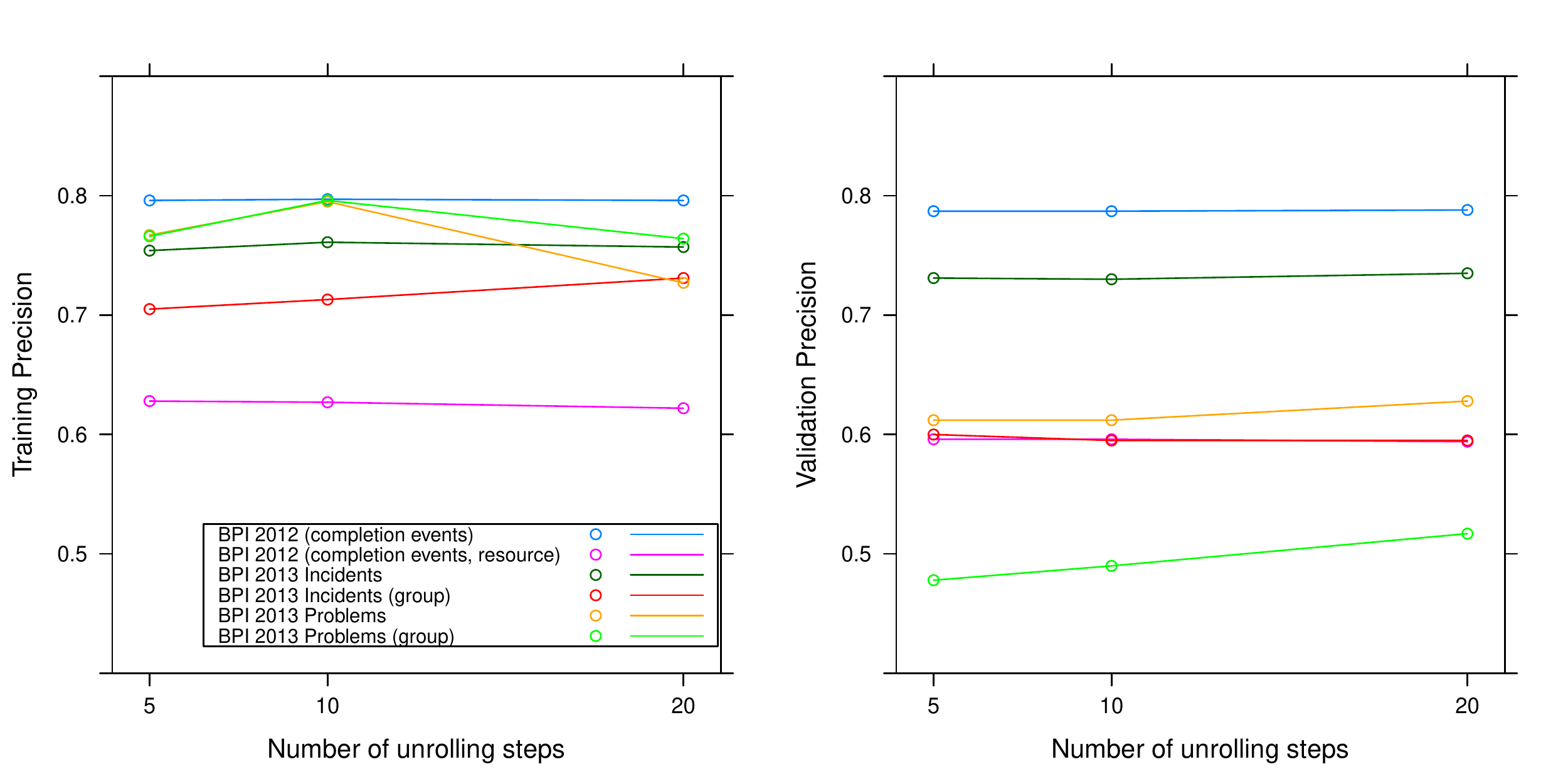}
\caption{Validation precision for three values of unrolled steps for selected datasets}
\label{fig:accuracybyunrollsteps}
\end{figure}

To identify the effect of the number of unrolled steps on prediction performance, we repeat our experiments using 10 and 5 unrolled steps, keeping the other parameters unchanged. Fig.~\ref{fig:accuracybyunrollsteps} and Table~\ref{tab:results.unrollsteps.validation} show the training and validation precision of selected datasets against the number of unrolled steps. Neither training nor validation precision is significantly affected by the number of unrolled steps. There appears to be a minor effect for the BPI 2013 Problem dataset (more pronounced when including the organizational group), where the training performance decreases with increasing number of unrolled steps, while the validation performance increases. This indicates that there are few to no long-term dependencies in the different processes that cannot be captured by even five previous steps, which is in line with the mutual information considerations in Section~\ref{sec:prediction}.

\begin{table}
\centering
\small

\begin{tabular}{|l|r|r|r||r|r|r|} \hline
\multirow{3}{*}{Dataset} & \multicolumn{3}{c||}{No Resource Info} & \multicolumn{3}{c|}{With Resource Info}  \\\cline{2-7}
& \multicolumn{3}{c||}{Unrolled Steps} & \multicolumn{3}{c|}{Unrolled Steps} \\
 & 20 & 10 & 5 & 20 & 10 & 5 \\ \hline

BPI2013.Incidents 						& .735 & .730 & .731 & .595 & .595 & .600 \\
BPI2013.Problems 						& .628 & .612 & .612 & .517 & .490 & .478 \\
BPI2012 (completion events) 				& .788 & .787 & .787 & .594 & .596 & .596 \\
BPI2012 (all events) 					& .859 & .860 & .860 & .663 & .667 & .667 \\
BPI2012.W (completion events) 			& .658 & .657 & .657 & .208 & .201 & .201 \\
BPI2012.W (all events) 					& .832 & .831 & .831 & .559 & .558 & .558 \\
BPI2012.A 							& .832 & .833 & .834 & .709 & .710 & .712 \\
BPI2012.O						 		& .836 & .836 & .836 & .590 & .590 & .590 \\
\hline

\end{tabular}
\caption{Validation precision for different number of unrolled steps}
\label{tab:results.unrollsteps.validation}
\end{table}

\subsection{Interpreting the RNN}

One of the main criticisms of neural networks is the fact that no explicit, human understandable model is created; the learning consists entirely in the optimization of the values of thousands or millions of floating point parameters. We present two ways in which users can gain insight into what the neural net has learned, i.e. the knowledge that is encoded in its structures.

\subsubsection{Process Hallucinations}

Neural networks, especially RNN, can be made to ''hallucinate'', i.e. to generate output on their own, for example in language modeling \citep{DBLP:journals/corr/Graves13,DBLP:conf/icml/SutskeverMH11,karpathyblog} or music creation \citep{DBLP:journals/corr/OordDZSVGKSK16, DBLP:journals/corr/HuangW16,DBLP:journals/corr/ChoiFS16}. This is done by feeding the net output (prediction) immediately back as new input. Hallucinations can provide insight into what the RNN has learned about the training set. The ability of an RNN to re-produce, on its own, realistic and convincing process traces demonstrates that it has correctly learned the important features of the event log used as training sample and thereby validates the trained net and supports the usefulness of the RNN approach for process prediction.

Hallucinations are initialized by providing a short seed sequence of input events. One may either accept the event with the highest probability as output, or one samples from events using the output probabilities. We explore the two methods for the full BPI 2012 and BPI 2013 datasets. For each output  method and for each dataset, we train a net with 32 embedding dimensions and 5 unrolled steps for 100 epochs. After training is complete, we produce 20 hallucinations of sequence length 1000 from each trained net. An excerpt of the BPI2012 hallucinations is shown in Fig.~\ref{fig:hallucination_example}. The complete set of generated hallucinations is available from the authors.

\begin{figure}
\footnotesize

\begin{framed}
\begin{ttfamily}
ASUBMITTED APARTLYSUBMITTED ADECLINED WAfhandelenleads [EOC] ASUBMITTED APARTLYSUBMITTED APREACCEPTED WAfhandelenleads WCompleterenaanvraag WCompleterenaanvraag WCompleterenaanvraag AACCEPTED AFINALIZED OSELECTED OCREATED OSENT WCompleterenaanvraag WNabellenoffertes WNabellenoffertes WNabellenoffertes OSENTBACK WNabellenoffertes WValiderenaanvraag WValiderenaanvraag WValiderenaanvraag WNabellenincompletedossiers OCANCELLED OSELECTED OCREATED OSENT WNabellenincompletedossiers WNabellenincompletedossiers WNabellenincompletedossiers WNabellenincompletedossiers WNabellenincompletedossiers ACANCELLED OCANCELLED WNabellenincompletedossiers [EOC] ASUBMITTED APARTLYSUBMITTED WAfhandelenleads ADECLINED WAfhandelenleads [EOC] ASUBMITTED APARTLYSUBMITTED ADECLINED [EOC] ASUBMITTED APARTLYSUBMITTED APREACCEPTED AACCEPTED OSELECTED AFINALIZED OCREATED OSENT WCompleterenaanvraag WNabellenoffertes WNabellenoffertes ADECLINED ODECLINED WNabellenoffertes [EOC]
\end{ttfamily}
\end{framed}

\caption{Example hallucinations for the BPI2012 dataset (probabilistic output sampling, ''[EOC]'' designates end-of-case)}
\label{fig:hallucination_example}
\end{figure}

In addition to visually inspecting the generated hallucinations and judging their realism, we mine a process model from the original event log using the Flexible Heuristics Miner (FHM) \citep{DBLP:conf/cidm/WeijtersR11} with default parameters. We choose the FHM because it performs well on a wide variety of models \citep{DBLP:conf/bpm/ClaesP12}. We replay both the original log and the generated hallucinations against this model and compute the replay fitness, which indicates how well the model can generate the observed behavior \citep{promreplay}. A replay fitness of the hallucinations similar to that of the original model is a strong indication that the generated traces are realistic. We also examine whether the frequency distribution of events in the generated hallucinations matches that in the original dataset, using the Kolmogorov-Smirnov test. A non-signifant test (p-value $> 0.05$) indicates the frequency distributions come from the same underlying population.

Table~\ref{tab:hallucinations} shows the results for the BPI2012 dataset. Probabilistic output sampling produces more realistic results than choosing the output with the highest probability; the latter produces many long uniform event sequences that are uncharacteristic of the input logs. Results for the other event logs are similar. In summary, the generated hallucinations show that the neural networks learn the relevant features of the input event logs. 

\begin{table}
\centering
\small

\begin{tabular}{|l|r|r|r|} \hline
                           &  full log & prob. sampling & top-p output \\ \hline
Mean trace length          & 13   & 12   & 21 \\
Max trace length           & 96   & 56   & 194 \\
Replay fitness      	  & .586 & .582 & .350 \\
KS test p-value		  & ---  & .878 & .000 \\
\hline
\end{tabular}
\caption{Hallucination characteristics for BPI2012 (completion events only) dataset}
\label{tab:hallucinations}
\end{table}

Hallucinations can also be used to predict the remainder of a case. For this, the hallucination is initialized with a trace prefix and then continued until it produces an end-of-case indicator. The hallucination can then be compared with the actual trace continuation using a string-edit distance. We train nets with 32 embedding dimensions and 5 unroll steps for 100 epochs. Using trace prefixes of length 5, we produce hallucinations and compare them to the actual continuation using the normed Damerau--Levenshtein distance, which ranges from 0 to 1 (Table~\ref{tab:remainder}).

\begin{table}
\centering
\small

\begin{tabular}{|l|r|r|} \hline
\multirow{2}{*}{Dataset} & \multicolumn{2}{c|}{Damerau--Levenshtein Distance} \\
 & Mean & SD \\ \hline
BPI2013.Incidents 				& \parbox{.8in}{\raggedleft .563} & .199 \\
BPI2013.Problems  				& .616 & .177 \\
BPI2012 (completion events) 		& .659 & .203 \\
BPI2012.W (completion events) 	& .703 & .205 \\
BPI2012.W (all events) 			& .697 & .211 \\
BPI2012.A 					& .545 & .241 \\
BPI2012.O 					& .532 & .191 \\
\hline
\end{tabular}
\caption{Mean and standard deviation of Damerau-Levenshtein distance between actual trace remainders and those predicted from a prefix of length 5. Hallucinations produced using probability sampling ($k=1$) and element-wise feedback ($m=1$); smaller is better. }
\label{tab:remainder}
\end{table}

The application of hallucinations might be interesting in other areas of process mining as well, such as improving event log completeness in cases where particular mining algorithms benefit from better or larger logs\footnote{We thank one of the anonymous reviewers for this suggestion.}.

\subsubsection{Hidden State Dynamics}

Recent work on understanding RNNs also focuses on visualization. In particular, visualizations of the embedding matrix, the state activation and the state dynamics are useful in understanding how an RNN encodes knowledge \citep{DBLP:journals/corr/KarpathyJL15,DBLP:journals/corr/LiCHJ15,DBLP:journals/corr/YosinskiCNFL15,DBLP:journals/corr/StrobeltGHPR16}. 

We export embedding matrices after completion of training to create 2D t-SNE plots \citep{maaten2008visualizing}. They are not included here as no significant or obvious clustering of events is discernible for any of our datasets. 

We use LSTMVis \citep{DBLP:journals/corr/StrobeltGHPR16}, which is motivated by earlier work by \citet{DBLP:journals/corr/KarpathyJL15}, to examine the activation of hidden state cells for different inputs\footnote{Because LSTMVis provides an interactive visualization for exploring state dynamics, it is difficult to provide in static form in this paper. The fully interactive visualization is available at \url{http://rnnprocess.evermann.ca}}. The visualization assists in identifying cells that are active for some input but not others \citep[the ''hypothesis selection process'',][]{DBLP:journals/corr/StrobeltGHPR16}, and then confirming such hypotheses by comparing activation patterns against similar input \citep[the ''matching process'',][]{DBLP:journals/corr/StrobeltGHPR16}. 

\begin{figure}
\centerline{\includegraphics[width=\textwidth]{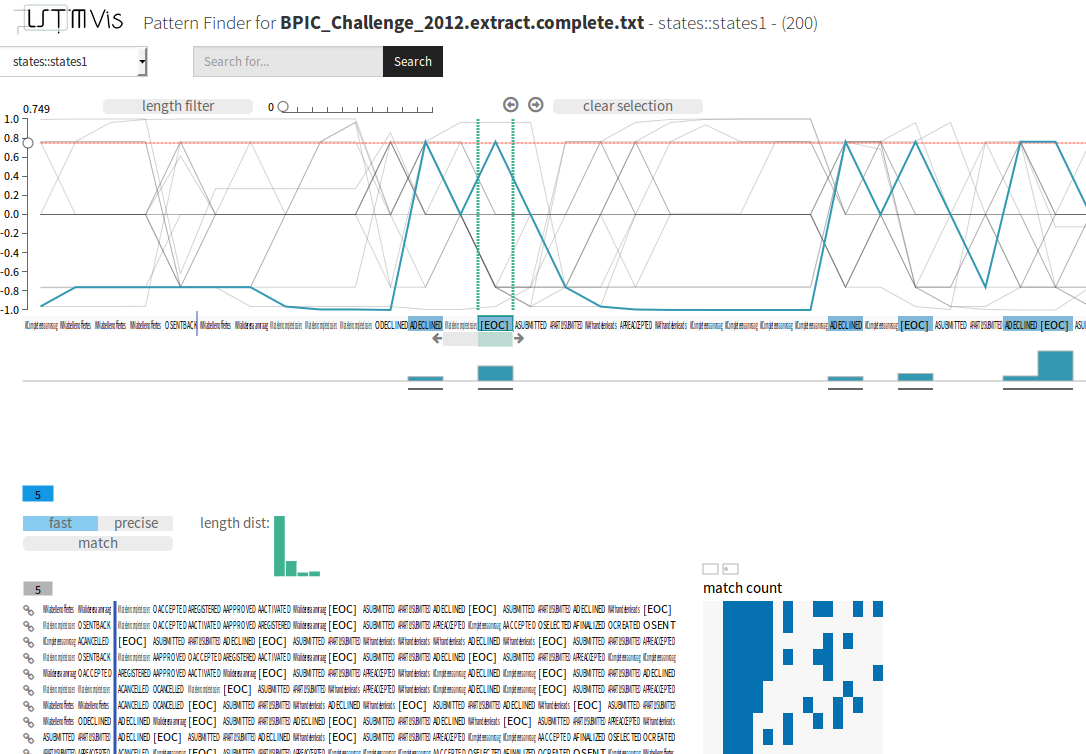}}
\caption{LSTM visualization of states for declined loan application processes}
\label{fig:lstmvis1}
\end{figure}

Figure~\ref{fig:lstmvis1} shows the tool in use with hidden states on the first level of the trained RNN for the BPI2012 dataset with 16 embedding dimensions. The event sequence ''ADECLINED [EOC]'' is selected in the timeline (top part of figure). The inputs that match the activation pattern of the selected event sequence are sequences of the same two events, shown at the bottom of the image. This cell appears to represent the ending of cases with declined loan applications. In some cases, another event occurs in between the ''ADECLINED'' and ''[EOC]'' and this input is represented by the same cell activation pattern. Other hidden state cells represent the events ''WNABELLENOFFERTES'', ''WCOMPLETERENANVRAAG'', and the sequence ''AACCEPTED AFINALIZED OSELECTED''. 

\section{Discussion and Conclusion}
\label{sec:discussion}

This paper introduced the use of deep learning for process prediction. Our approach does not rely on explicit process models and can be applied when models do not exist or are difficult to obtain. Our results, surpassing or close to the state-of-the-art and with cross-validated precision in excess of 80\% on many problems, demonstrate the feasibility and usefulness of this approach. 

While one can perform prediction by mining a model from event logs, and mining decision rules for each process branch point \citep{DBLP:journals/kais/LakshmananSDUK15,DBLP:journals/kais/UnuvarLD16}, this has inherent drawbacks. Process mining algorithms trade off different quality criteria, such as fitness, precision, generalizability, and simplicity. The mined models are abstractions and intentionally imperfect representations of the underlying true process. This impairs predictive performance. Decision mining at process branch points or for trace prefixes \citep{DBLP:conf/dis/CeciLFCM14} makes trade-offs between recall, precision, and parsimony, and may use simplifying assumptions such as linearity. This further impairs predictive performance. Explicit process models and decision rules are useful for understandability. At the same time, however, their intentional abstraction and parsimony makes them less suitable for prediction. In contrast, deep learning networks with their larger parameter space and non-linearity ''skip'' the intentionally imperfect representations between raw event logs and event prediction. 

We chose RNNs for their natural fit to the sequence data of process event logs. An alternative to RNN are n-gram models, in which a non-recurrent neural network is trained on all fixed-length trace prefixes of length $n$. The RNN approach has the advantage of allowing prediction from trace prefixes of arbitrary length while also offering better prediction performance \citep{mikolov2011extensions}. Alternatives to neural networks are probabilistic automatons such as HMM \citep{DBLP:journals/kais/LakshmananSDUK15,DBLP:journals/kais/UnuvarLD16} and PFA \citep{Breukeretal:2016}. Section~\ref{sec:prediction} showed that the event logs chosen for this study have characteristics that make either of these techniques applicable. Our prediction results confirm this. While our RNN are not always superior, they surpass the performance of PFA in many cases. However, other processes may involve long-distance dependencies between events that are not found in our data. Such long-distance dependencies in an event log are likely to present problems for HMMs and n-gram models, but not for RNN \citep{2016arXiv160606737L}.

Deep learning approaches are susceptible to overfitting. The larger a neural network, the better it can capitalize on chance training data idiosyncrasies. With an embedding space of 500 dimensions and a small event log, \citet{Evermannetal:2016} report training data prediction precision in excess of 99\%, showing clear overfitting. In this paper, we have also seen overfitting with similarly large embedding spaces. In the context of process model discovery, overfitting has its analogue in mined models with high fitness and precision, that are complex and not generalizable. And similar to quality trade-offs in process model mining, deep learning also makes trade-offs. Prediction precision is traded off against generalizability (by overfitting) and simplicity (network size, embedding dimensionality). Just as model fitness must be evaluated in the context of other quality dimensions, predictive performance must be evaluated in the context of generalizability and model size. An assessment of overfitting through cross-validation is therefore essential.

Our approach constructs the RNN inputs by concatenating categorical, character string valued event attributes and then encodeing these attributes via an embedding space. This is feasible only because of the small number of unique values for each attribute in our datasets. Because of this, the number of unique values of the concatenated input remains small. Hence, the embedding space dimensionality need not be large. While the names of executing activities and lifecycle transitions for most datasets usually have a small set of values, the set of resources or organizational groups may be much larger, which limits the feasibility of this approach. Application-specific event-level or case-level attributes can also be included in predictors and predictands simply by concatenating their values. This too is limited to cases where the set of unique values is small. Numerical attributes are difficult to handle in our approach unless they are encoded in a limited number of intervals, which may lead to significant information loss. 

A different approach, first encoding individual attributes and subsequently concatenating the embedding vectors for input to the RNN, can keep the RNN input size small even when the number of unique values for some attributes is large, and also simplifies including numerical attributes. Exploring this alternative and its predictive performance, and determining ways to identify event- and case-level attributes that are useful as predictor variables, are important future work.

\singlespacing

\section*{Acknowledgements}

The authors gratefully acknowledge the support of the Memorial University Center for Health Informatics and Analytics, St. John's, Canada and the Hasso-Plattner-Institute at the University of Potsdam, Germany, in providing access to computing resources.

\section*{References}

\singlespacing 

\bibliographystyle{elsarticle-harv}

\end{document}